\pdfoutput=1
\documentclass{article}
\usepackage{soul}
\usepackage{lineno,hyperref}
\usepackage{multirow}
\usepackage[final,nonatbib]{nips_2016}
\usepackage[numbers]{natbib}

\usepackage[utf8]{inputenc} 
\usepackage[T1]{fontenc}    
\usepackage{hyperref}       
\usepackage{url}            
\usepackage{booktabs}             
\usepackage{nicefrac}       
\usepackage{microtype}      
\usepackage[pdftex]{graphicx}

\usepackage{mathptmx}
\usepackage[cmex10]{amsmath}
\usepackage{amsfonts}

\usepackage{array}
\newcolumntype{L}[1]{>{\raggedright\let\newline\\\arraybackslash\hspace{0pt}}m{#1}}
\newcolumntype{C}[1]{>{\centering\let\newline\\\arraybackslash\hspace{0pt}}m{#1}}
\newcolumntype{R}[1]{>{\raggedleft\let\newline\\\arraybackslash\hspace{0pt}}m{#1}}

\newcommand{\units}[1]{\,\text{#1}}
\newcommand{\nuclearSym}[2]{${{}^{#1}\text{#2}}$}

\title{Automated X-ray Image Analysis for Cargo Security: \\Critical Review and Future Promise}

\author{
  Thomas W.~Rogers$^{1,3}$ \quad Nicolas~Jaccard$^{1}$ \quad Edward J.~Morton$^{2}$  \quad Lewis D.~Griffin$^{1*}$\\\\
  $^1$Department of Computer Science, University College London, London, UK\\
  $^2$Rapiscan Systems Ltd., Stroke-on-Trent, UK\\
  $^3$Department of Security and Crime Sciences, University College London, London, UK\\\\
  \small{$^{*}$Correspondence: l.griffin@cs.ucl.ac.uk}
}

\begin{document}

\maketitle

\begin{abstract}
We review the relatively immature field of automated image analysis for X-ray cargo imagery. There is increasing demand for automated analysis methods that can assist in the inspection and selection of containers, due to the ever-growing volumes of traded cargo and the increasing concerns that customs- and security-related threats are being smuggled across borders by organised crime and terrorist networks. We split the field into the classical pipeline of image preprocessing and image understanding. Preprocessing includes: image manipulation; quality improvement; Threat Image Projection (TIP); and material discrimination and segmentation. Image understanding includes: Automated Threat Detection (ATD); and Automated Contents Verification (ACV). We identify several gaps in the literature that need to be addressed and propose ideas for future research. Where the current literature is sparse we borrow from the single-view, multi-view, and CT X-ray baggage domains, which have some characteristics in common with X-ray cargo.
\end{abstract}

\section{Introduction}
\label{sec:introduction}
The use of cargo containers in global trade transactions continues to grow. From 2004 to 2014, and despite the 2008 global economic crisis, the number of Twenty-foot Equivalent Unit (TEU) container transactions more than doubled to reach almost $7{\times}10^8$ TEU per annum~\cite{worldBank2016}. During this time, the US Container Security Initiative (CSI), proposed in the wake of the 9/11 terrorist attacks, has encouraged 100\% screening of containers~\cite{Romero2003}, and is being implemented by ports around the world~\cite{CSI2011}. With the ever-growing numbers of containers and increasingly stringent screening requirements, there has been active research in academia and industry to engineer accurate and rapid screening methods, which are vital for both the global economy and security.

Cargo containers are frequently exploited for smuggling, which can be achieved by concealment amongst and within legitimate cargo or packaging, by concealment within legitimate or false container partitions, or by intercepting containers to plant and recover contraband (rip on/rip off)~\cite{EuCCh6}. Smuggling bypasses customs controls, allowing criminals to: avoid duties on legitimate goods (e.g. cars, alcohol, cigarettes); trade prohibited or counterfeit items; launder money; and avoid sanctions~\cite{EUCCh4}. 

Under the CSI and similar initiatives, cargo inspection is performed in three layers. The first layer selects samples of containers for inspection~\cite{CSI2011,EuCCh1}. Containers are selected based on specific intelligence or a risk analysis. Often, a small fraction of containers are randomly sampled in the hope of catching out criminals who have discovered ways to make shipments appear ``low risk''~\cite{EuCCh1}. Selected containers first undergo Non-Intrusive Inspection (NII). If anything suspicious is detected then the container is sent for physical inspection. Physical inspection is very slow and expensive; it has to be well documented for use as evidence and done carefully to avoid compensation payouts if the container is innocuous.

\begin{figure}[htbp]
\centering
\includegraphics[width=\textwidth]{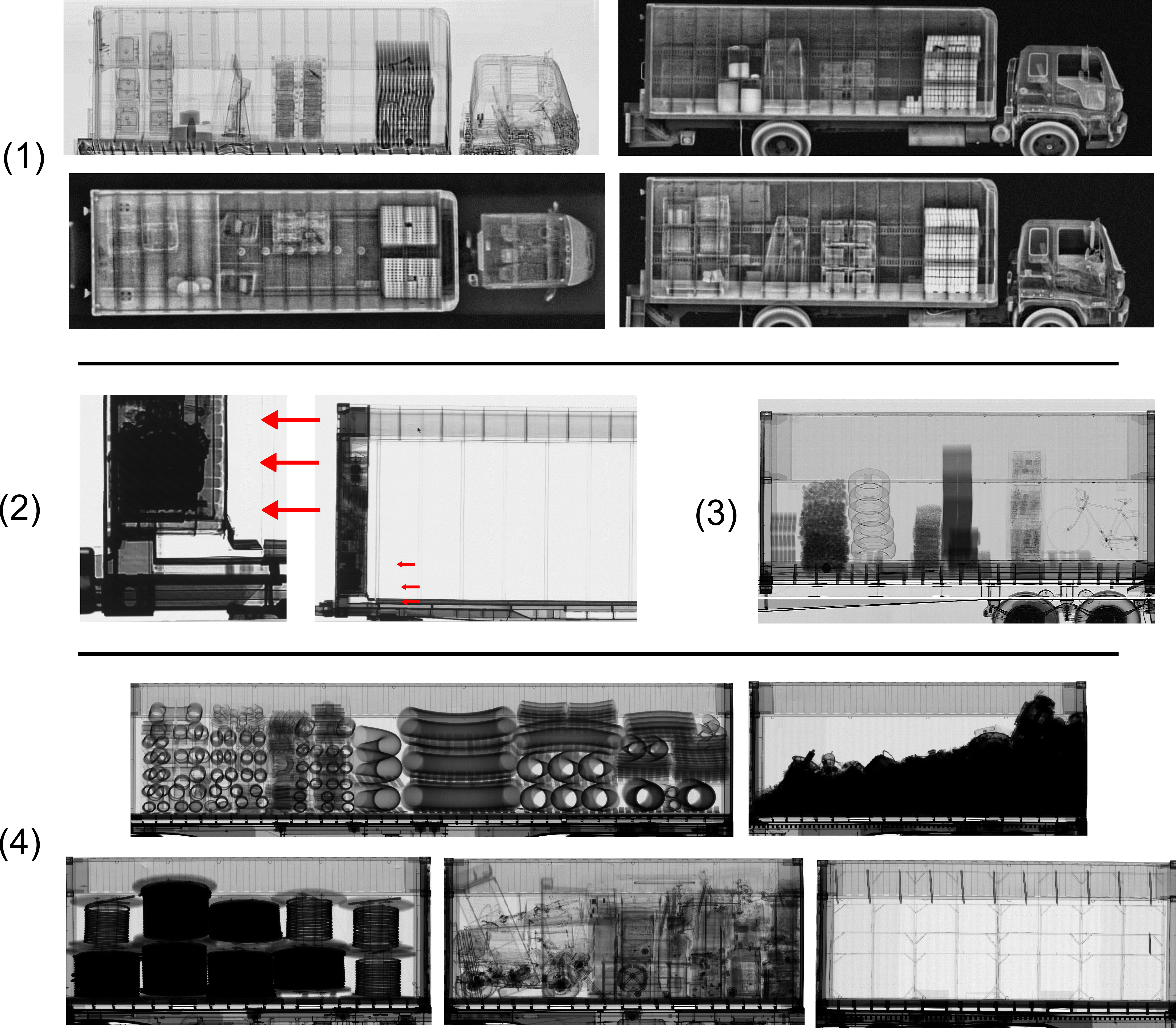}
\caption{Examples of cargo X-ray images, including: (1) triple-view backscatter and single transmission images from an AS\&E OmniView\textregistered\ Gantry; (2) example of concealed cigarettes (indicated by red arrows) in a transmission X-ray image of refrigerated container; (3) transmission image of 20 ft container carried by lorry; and (4) examples of transmission images from the Stream-of-Commerce (SoC) captured by a Rapiscan Eagle\textregistered\ R60 railscanner. Sources: Zheng and Elmaghraby~\cite{Zheng2013a}; Vogel~\cite{vogel2007vehicles}; Chalmers \emph{et al.}~\cite{Chalmers2007}; and Rapiscan Systems.}
\label{fig:exampleImages}
\end{figure}

\begin{figure}[htbp]
\centering
\includegraphics[width=\textwidth]{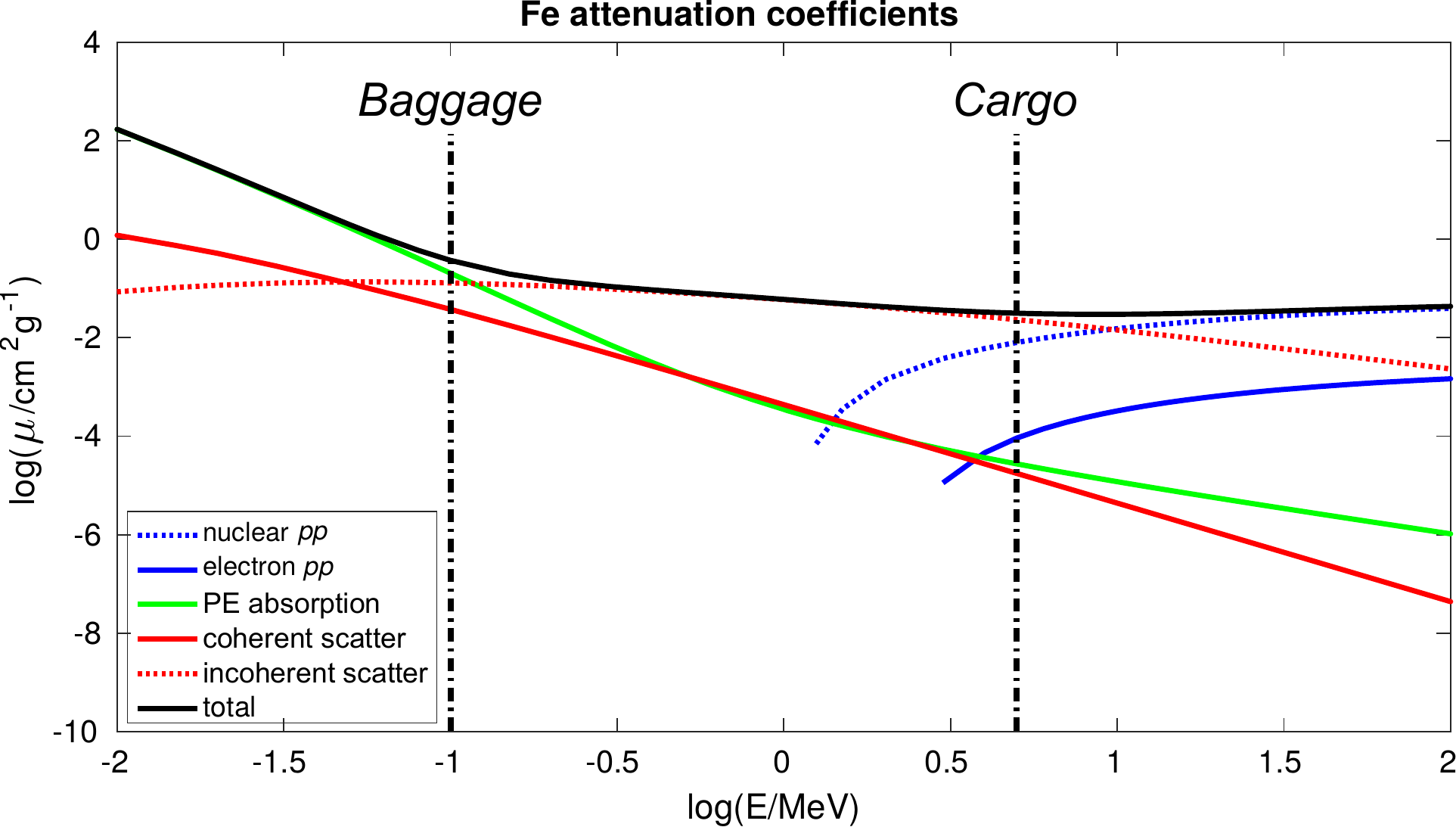}
\caption{Attenuation coefficient ($\mu$) contributions plotted for iron as a function of photon energy $E$. The vertical dashed lines indicate typical photon energies used in baggage and cargo screening. Estimation of $\mu$, which can be determined by the difference between images captured at two energies, is more difficult at cargo energies since the total $\mu$ gradient becomes small and large-scale commercial systems tend to suffer from severe noise. Data source: NIST~\cite{NIST}.}
\label{fig:attenuation}
\end{figure}

\begin{figure}[htbp]
\centering
\includegraphics[width=\textwidth]{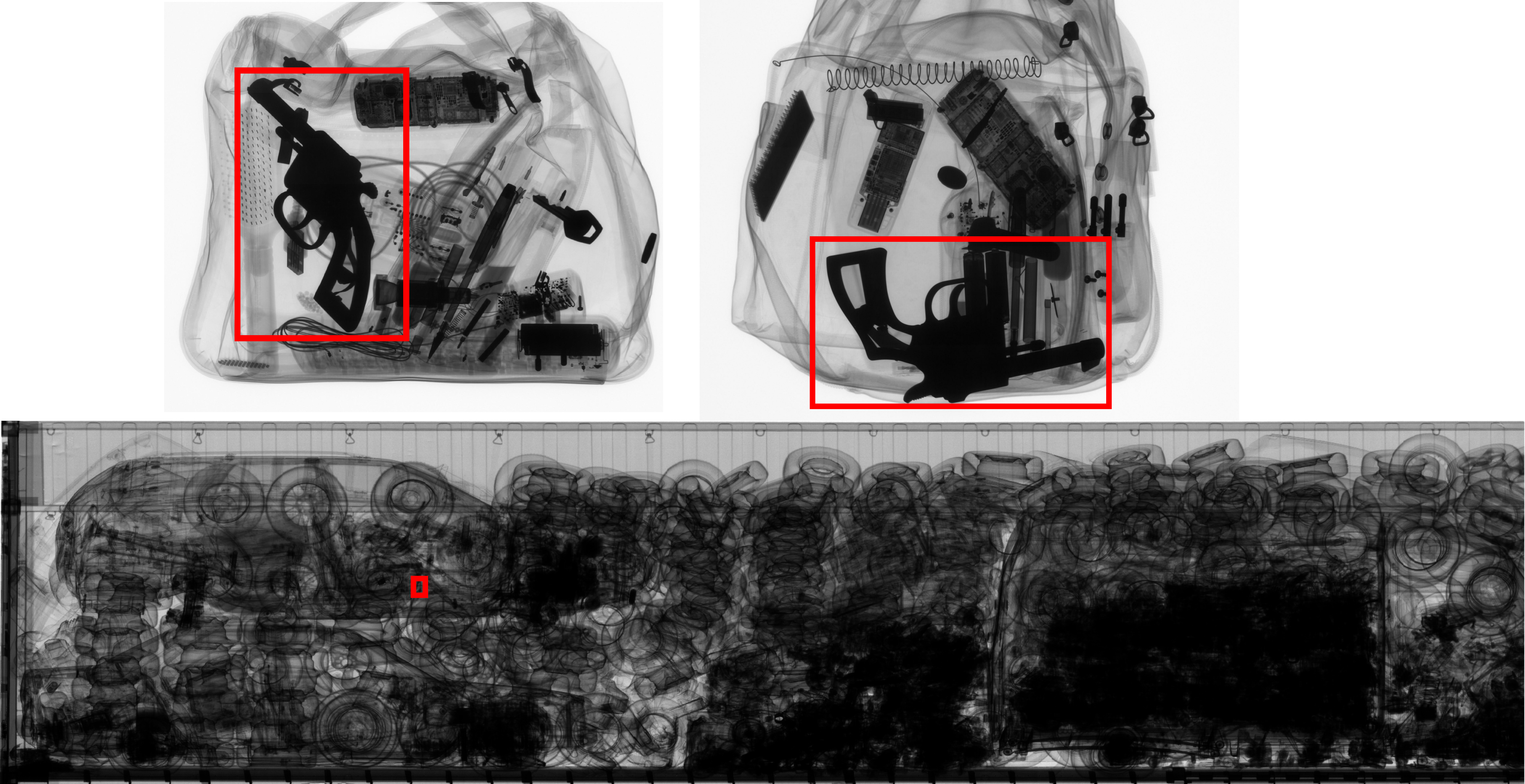}
\caption{Comparison of single energy transmission X-ray images of baggage (top) and cargo (bottom). The red boxes indicate the approximate size of a handgun in the image. In baggage, the handgun occupies a large portion of the image, and there is little distracting clutter, so that there is a strong visual signature. In cargo, there is typically a lot more clutter and the handgun occupies a very small portion of the image, make handgun detection a difficult for humans and algorithms. Sources: $\mathbb{GDX}\text{Ray}$ database~\cite{Mery2015} and Rapiscan Systems.}
\label{fig:cargovsbaggage}
\end{figure}

The majority of cargo NII systems use transmission X-ray or $\gamma$-ray radiography~\cite{Liu2008} to form an image of the cargo contents (examples in Fig.~\ref{fig:exampleImages}). The image is sent to a human operator who searches it for any anomalies, specific threats, or discrepancies with the shipping manifest. Cargo images pose a difficult visual search task for the human operator, and they are much more difficult to analyse than other types of border security imagery such as baggage. This is because cargo scanners have to operate at a much larger scale. For example, a 40\units{ft} General Purpose cargo container has a volume of $67.6\,\text{m}^3$~\cite{EUCCh3} and is made out of steel, whereas hand luggage volume\footnote{Determined based on British Airways cabin bag size allowance of $56\units{cm}{\times}45{\times}\units{cm}{\times} 25\units{cm}$} is typically $0.063\,\text{m}^3$ and usually made out of fabric or plastics. The physical scale of cargo scanners makes it difficult to efficiently perform 3D Computed Tomography (CT)~\cite{calvert2013preliminary} but some multi-view systems do exist. Moreover, for cargo it is more difficult to extract material composition information due to the higher energies required for sufficient penetration to obtain good image contrast (Fig.~\ref{fig:attenuation}). Cargo images are also far more cluttered, whilst small threats, such as firearms, have a very small visual signature. A comparison between baggage and cargo single-view X-ray imagery is shown in in Fig.~\ref{fig:cargovsbaggage}.

\begin{figure}[htbp]
\centering
\includegraphics[width=\textwidth]{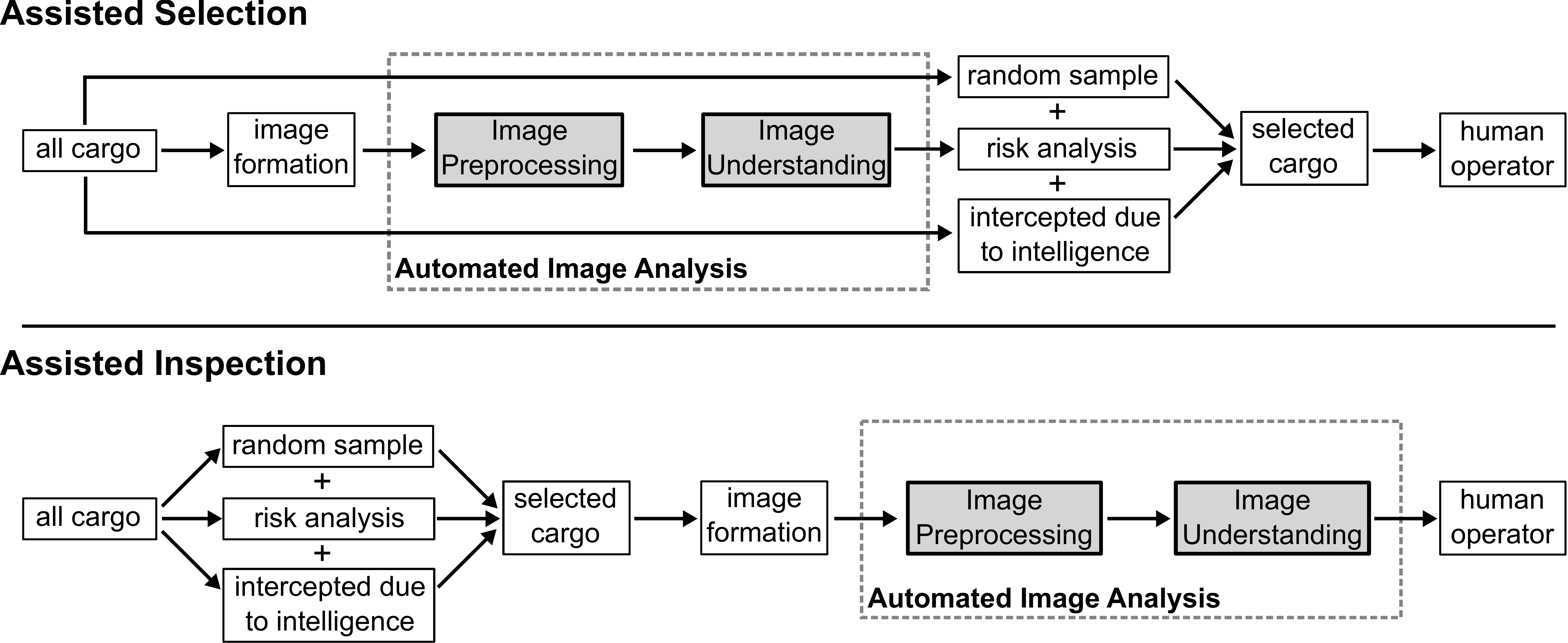}
\caption{Process diagrams showing the cargo inspection process and possible uses of automated image analysis; Assisted Selection (top) and Assisted Inspection (bottom). In Assisted Selection, image analysis is used to inform the risk analysis so that cargo is selected for inspection with greater accuracy, and thus reducing the burden on human operators. In Assisted Inspection, annotations (e.g. bounding boxes with a labelled confidence score) determined in the Image Understanding step are added to the image so that the operator can more quickly identify threats.}
\label{fig:overview}
\end{figure}

Automated image analysis can help with cargo screening by Assisted Inspection or Assisted Selection (Fig.~\ref{fig:overview}). Currently most research has been geared towards Assisted Inspection, with algorithms designed to assist the operator, such as by annotating the image with a Region-of-Interest (ROI) to prompt the operator of a potential security- or customs-related threat. The goal of Assisted Selection is to use automated image analysis to inform the risk analysis used for cargo selection, but relies on the ability to scan all containers at high throughput rates. Such technologies are becoming available, such as rail scanners capable of imaging cargo travelling at up to 60\units{km/h}~\cite{R60}. When such systems are widely deployed, Assisted Selection has the potential to increase true positive and reduce false positive cargoes in the selected sample. In doing so, it should allow for human resources to be allocated more efficiently.

The literature on automated image analysis for cargo can be separated into Image Preprocessing, and Image Understanding. Image Preprocessing is a broad category including any operation made to an image in order to help Image Understanding by either humans or algorithms. Image Preprocessing includes: image manipulation; image correction and denoising; material discrimination and segmentation; and Threat Image Projection (TIP). Image Understanding is about decisions that are made based on the image contents. Currently, the literature is split into Automated Threat Detection (ATD) and Automated Contents Verification (ACV).

In this paper we investigate the current literature according to the themes of Image Preprocessing (Sec.~\ref{sec:imagepreprocessing}) and Image Understanding (Sec.~\ref{sec:imageUnderstanding}). In some cases, the literature directly relating to cargo imagery is scarce. This is due largely to commercial and security protection, and the difficulty for academics to obtain access to commercial scanning hardware. Additionally, the majority of funding goes towards aviation security, where search tasks are more tractable, and there is a more obvious and immediate threat from terrorism. In cases where cargo research is sparse, we look to the literature from other domains such as baggage, since many of the findings there may be transferable to the cargo domain. The purpose of this paper it to map out the current literature, to identify gaps in it, and to propose future directions of research.

\section{Image Preprocessing}
\label{sec:imagepreprocessing}
We define Image Preprocessing as any process which is performed before, and in order to improve the performance of, human or automated Image Understanding. In the literature, we have identified four topics: image manipulation; image quality improvement; material discrimination and segmentation; and Threat Image Projection (TIP).

\subsection{Image manipulation}
\label{subsec:imageManipulation}
\begin{figure}[htbp]
\centering
\includegraphics[width=1\textwidth]{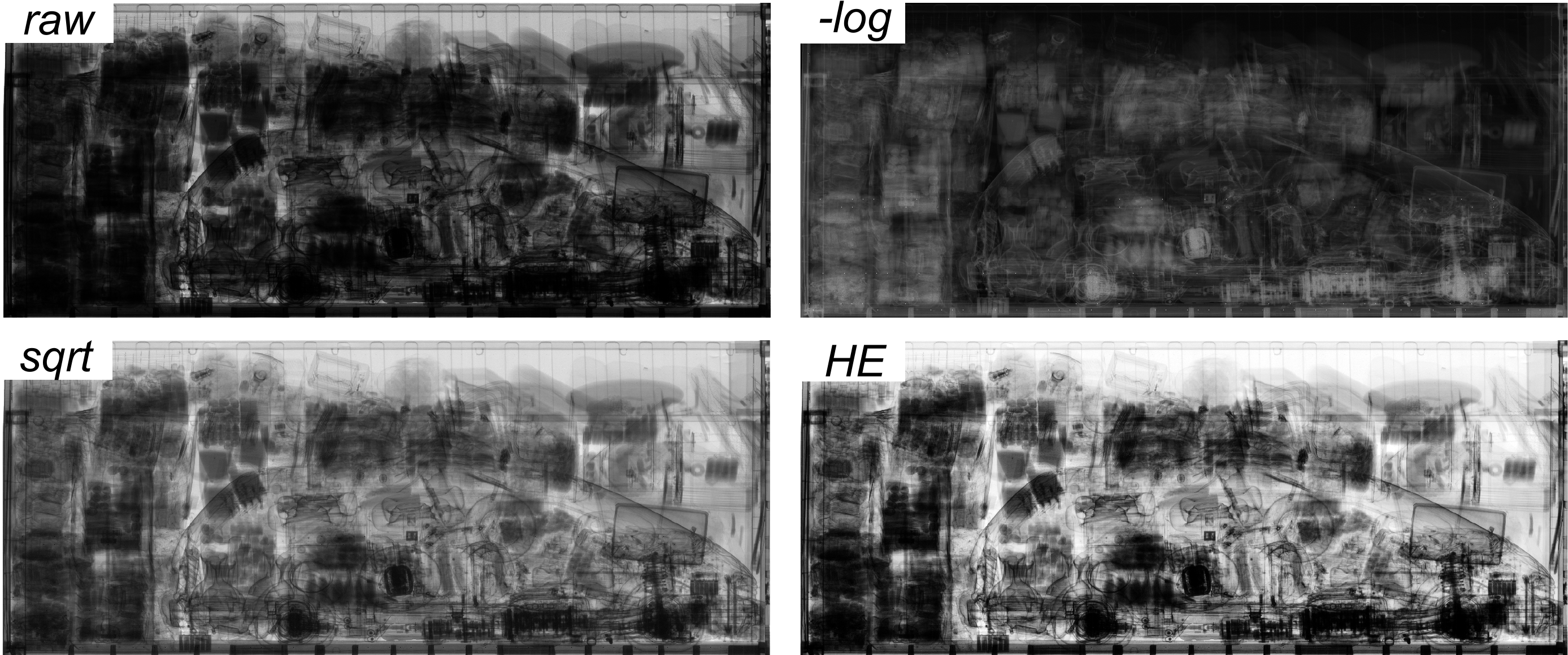}
\caption{Examples of different image manipulations for an X-ray image of a car with partial occlusion. Clockwise from top-left: raw high energy image; negative logarithm of the image with pixel values proportional to the material attenuation coefficient and thickness; square-root of the image; and Histogram Equalisation (HE) so that the image histogram is flat. There is still some debate over which manipulation allows best human or algorithmic Image Understanding.}
\label{fig:manipulations}
\end{figure}

Image manipulation is used to improve the accuracy of human operators and automated Image Understanding algorithms. Most work has been on studying the threat detection performance of human operators under different image manipulation functions implemented in commercial image viewing software. Manipulations include pseudocolouring, edge enhancement, and intensity transforms such as Histogram Equalization (HE), logarithm and square-root (Fig.~\ref{fig:manipulations}). Note that pseudocolour is not based on material properties, which we discuss in Sec.~\ref{subsec:MatSep}.

For cargo screening, Michel \emph{et al.}~\cite{Michel2014a} have shown that image pseudocolouring does not offer improved performance over the raw greyscale image, when identifying narcotics, weapons, Improvised Explosive Devices (IEDs) and other explosives. Similar results have been found by Klock~\cite{Klock2005}, who tested human performance at detecting concealed IEDs, guns, knives and other prohibited items in baggage. Evaluated manipulations included pseudocolour, greyscale, inverse, inorganic or organic material stripping, and a commercial Crystal Clear\texttrademark\ function\footnote{Details of Crystal Clear\texttrademark\ are difficult to find, but the function ``optimises image contrast and resolution to bring out picture details'' according to a public verbal communication by Andreas Kotowski (Rapiscan Systems CTO) in 2001.}. They found that greyscale and the Crystal Clear\texttrademark\ functions best aided human performance.

Chen~\cite{Chen2005} reasons that although most X-ray cargo images are captured and encoded in 16\units{bits}, typical greyscale displays only use 8\units{bits} so that useful information is lost, but with pseudocoloured images, there are 8\units{bits} available each for each colour channel, thus preserving the information. However, he argues that the effectiveness of pseudocolour is in fact limited by the ability of humans to detect subtle colour differences. The author also claims that edge enhancement techniques do not work well for cargo due to the complexity of objects and high pixel noise. Chen~\cite{Chen2005} qualitatively evaluates linear, logarithm and Adaptive HE (AHE) image transforms. He argues that log transform can be beneficial as it makes image brightness proportional to object thickness, but thin items are sometimes lost. The square-root transform can be beneficial since the Signal-to-Noise ratio (SNR) is proportional to the square-root of pixel intensity, thus it is an equal-noise display method. Finally, the author observes that, qualitatively, AHE is the best method but that full object thickness information is lost.

As far as we are aware, there have been no specific studies on the effect of image manipulations on automated cargo Image Understanding. A few researchers have done small studies as parts of larger bodies of work. For example, when building a car detector, Jaccard \emph{et al.}~\cite{Jaccard2014} tested log-intensity histograms as a feature and found them to perform better than intensity histograms. However, the log-image gave worse performance than the raw image when using oriented Basic Image Features (oBIFs), possibly because the oBIF parameters were not re-tuned. In a later paper~\cite{JaccardCars16}, this time trying both hand-crafted Pyramid Histograms of Visual Words (PHOW) features and features learnt using trained-from-scratch Convolutional Neural Networks (CNNs), the authors found that using log transformed images as input gave a substantial improvement in performance.

Other researchers have applied different image manipulations before applying Image Understanding algorithms. These include: Gaussian blurring~\cite{Zhang2014}; rudimentary segmentation algorithms to extract different image regions~\cite{Zhang2014,Rogers2015}; and image inversion followed by z-score normalisation and Retinex filtering~\cite{Zheng2013a}.

\subsection{Image quality improvement}
\label{subsec:imageQualityImprovement}
Image quality improvement can include denoising methods to ameliorate Poisson or salt-and-pepper noise, and methods to correct image errors that arise during image acquisition.

To our knowledge, there have been no published comparison studies on different cargo image denoising techniques. However, in baggage, Mouton \emph{et al.}~\cite{Mouton2013} perform a comparative study on a number of denoising techniques applied to low quality baggage imagery. Techniques included: anisotropic diffusion; Total Variation (TV) denoising; bilateral filtering, translation-invariant wavelet shrinkage; Non-Local Means (NLM) filtering; and Alpha-Weighted Mean Separation and Histogram Equalisation (AWMSHE). They assess performance by running a Scale-Invariant Feature Transform (SIFT) point detector across image before and after the denoising. They identify object feature points (located on an object of interest) and noise (not on the object) within the CT image (i.e. assumed to be caused by noise or artefacts). Performance is then measured by taking number of object feature points as a fraction of the total number of feature points, assuming that an increasing ratio is indicative of improved performance using a SIFT-based detection algorithm. They find that all methods offer improved performance, over using just the raw image, with translation-invariant wavelet shrinkage performing best. However, it is unclear whether these results would generalise to algorithms that are not based on SIFT.

To our knowledge, there is one publication on image error correction for cargo. Rogers \emph{et al.}~\cite{Rogers2014} propose a method for correcting wobble artefacts in images captured by mobile cargo scanners. The wobble artefact originates from the wobble of the detector array as a mobile scanner traverses a stationary cargo. The method relies on a slight modification to the scanning hardware by rotating four of the imaging detectors by $90^\circ$ so that they can measure the beam across its width. This allows the beam to be tracked as it jitters on the detector array. The position tracking is achieved by fitting a Gaussian model to the beam cross-section to obtain an instantaneous estimate of the beam centroid. The instantaneous estimate is Bayesian fused with a second estimate of the beam position, which is a linear combination of previous estimates (i.e. auto-regression). This method improves the tracking robustness to heavily attenuating objects that obscure the beam. The authors use the beam position estimates to apply image corrections. They determine that they can fix: $70\%$ of image error due to detector wobble; $68\%$ of noise due to source fluctuation; and $95\%$ of noise due to sensor variation.

Other researchers have applied simple image correction and denoising methods in preprocessing prior to algorithmic Image Understanding. These include: median filtering or filling individual erroneous pixels their neighbourhood median to remove salt-and-pepper noise~\cite{Rogers2015,Jaccard2014}; normalisation of image columns to reduce errors from X-ray source fluctuation~\cite{Rogers2014,Rogers2015}; deletion of image rows or columns that contain no image information due to source miss-fire or detector downtime~\cite{Rogers2015,Jaccard2014}.

\subsection{Threat Image Projection (TIP)}
\label{subsec:TIP}
Threat Image Projection (TIP) is a technique first developed for baggage~\cite{Mitckes2003}. Most TIP methods insert a fictional threat from a database into an existing benign image. This can be used for Computer-Based Training (CBT) of operators, assessing their performance~\cite{Steiner-Koller2009}, or improving their detection performance by increasing their exposure rare threat scenarios~\cite{Godwin2010a}. Similarly, in cargo, researchers are exploring how TIP imagery can be used to increase the competency of operators, however, so far they have relied on screening experts manually merging \emph{threat} and \emph{innocuous} images using X-ray image merging software~\cite{Michel2014a}. Moreover, some researchers are beginning to use TIP as a data augmentation methodology when training Machine Learning (ML) based ATD algorithms.

In CT baggage, TIP is complicated by the 3D nature of images. TIP algorithms typically search for realistic placement volumes (voids)~\cite{Megherbi2012a,Yildiz2008} so that the projected threat does not intersect other objects and act as a visual cue for operators. Researchers have defined metrics for View Difficulty, Superposition and Bag Complexity~\cite{Schwaninger2007,Schwaninger2005,Schwaninger2004}. Such metrics can be used for adaptive CBT algorithms, where the difficulty of a given search task can be controlled. For example, if an operator is poor at finding threat items in certain contexts, such as complicated clutter, the algorithm can present more of these examples to improve performance under those contexts. Other researchers have realised that TIP imagery appears unrealistic, unless they generate realistic noise and artefacts that match those of the other objects in the baggage image. For example, Megherbi \emph{et al.}~\cite{Megherbi2012a,Megherbi2013} generate realistic metal artefacts in CT baggage, to ensure that artefacts in the threat are consistent with those in the rest of the baggage, and are not a visual cue for operators. Similar ideas are likely to be useful in cargo TIP, for example ensuring that magnification, pixel noise and scatter point-spread functions are consistent between the threat and the rest of the image.

In cargo, some authors have suggested methods for image synthesis for other purposes, but which could be applicable to TIP. White \emph{et al.}~\cite{White2008} introduce a method for generating synthetic $\gamma$-ray cargo radiographs, and use it as surrogate data for testing the effectiveness of different scanning systems when it is impractical to collect large amounts of empirical data. The authors derive an empirical model of the imaging system response from real radiographs of well-characterised objects. They claim to incorporate system properties such as sensitivity, spatial resolution, contrast and noise. To synthesise a threat image, the authors simulate photon transmission using a commercial ray-tracing package, and then apply smoothing and Gaussian noise consistent with their empirical measurements. The ray-tracing software allows for simulation of complex-object models, such as those developed in Computer-Aided Design (CAD). After simulating the photon transport and detector-response model the synthetic threat radiographs are injected into real radiographs. They perform this injection by pixel-wise multiplication of the synthetic threat radiograph with the real radiograph. This method comes directly from the Beer-Lambert law and assumes no cross-pixel effects such as scatter. We feel that synthesising threat images from 3D threat models, could prove invaluable in the future, particularly for adding emerging threats to TIP libraries, for example CAD models of 3D-printed weapons.

In the ML community, training data augmentation is used to improve the performance of ML-based algorithms. Data augmentation reduces overfitting by using label-preserving transformations to artificially enlarge the dataset~\cite{Krizhevsky2012}. Transformations must be realistic for the given imaging system, in order to make algorithms robust to natural variation. In visible spectrum imagery, examples of transformations include random crops (translation invariance), random flips (reflection invariance), and random addition of lighting (invariance to illuminance)~\cite{Howard2013}. In cargo X-ray imagery transformations could include variations in dose, perspective, material composition, and object orientation. Data augmentation is particularly useful in representation learning, such as deep CNNs, which are prone to overfitting if datasets are limited in size and variety. 

For automated cargo Image Understanding, researchers often face the problem of unbalanced datasets. Whilst images of non-threat cargoes are abundant, images of threat cargo are usually very rare in the wild. So researchers often rely on capturing staged threat images. This process is time consuming and expensive. Recently, researchers have begun to use TIP frameworks to help train and test ML-based Image Understanding algorithms. For example, they project staged threat images into innocuous Stream-of-Commerce (SoC) images, whilst adding realistic variation, to bring balance between the threat and non-threat classes. 

Rogers \emph{et al.}~\cite{Rogers2015} use image synthesis to train a classifier for detecting loads in declared-as-empty containers. The method extracts a database of objects from real cargo X-ray radiographs. An estimate of the background is obtained by exploiting the uniformity of the cargo container in the image vertical. Background removal is achieved by pixel-wise division of the cropped object by the background estimate. Extracted objects are then manipulated to create diversity in the training set. The authors include random object variations in translations, orientation, density, and volume. They also combine multiple random objects to form a composite object. The composite object is projected into real empty container images in a similar way to White \emph{et al.}~\cite{White2008}.

Jaccard \emph{et al.}~\cite{Jaccard2015} follow a similar process to Rogers \emph{et al.}~\cite{Rogers2015}, but for training a CNN from scratch to detect Small Metallic Threats (SMTs). They create very large numbers of threat images, with high variability in background appearances, by injecting threat radiographs into real radiographs by multiplication. They achieve background removal by manually delineating the threat item and background clutter, and dividing by the mean of the non-clutter background. To increase threat variability, the authors randomly position and flip the object, whilst varying the threat attenuation by a random factor between $0.95$ and $1.05$. The authors sample a total of $1.2{\times}10^4$ threat backgrounds from a very large number of real cargo radiographs.

Recently, Jaccard \emph{et al.}~\cite{Jaccard2016} have introduced a TIP module useful for training classifiers. They list a number of methods for adding realistic noise and variation to training data. This includes: object volume scaling by jointly scaling the in-plane area and the object attenuation; object density scaling by scaling the object attenuation; object flips, formation of composite threat objects; addition of noise; and varying the background appearance. More recently Rogers \emph{et al.}~\cite{rogers2016threat} have introduced a method for magnifying the object according to the depth of the object in the scene. Since most X-ray scanner use a divergent fan-beam the object appears taller as it is moved closer to the source. They suggest generating the vertical scale factor $\alpha$ by 
\begin{equation}
\alpha = 1 + d\left(\frac{l_1}{l_0}-1\right),
\end{equation} 
where $d\in[0,1]$ is the normalised depth position of the object from the source, and $l_0$ and $l_1$ are the vertical lengths of an object placed at $d=0$ and $d=1$, respectively. The authors also suggest adding Poisson and salt-and-pepper noise to images, and briefly propose a method for adding illumination variation due to detector wobble in mobile systems.

\subsection{Material discrimination and segmentation}
\label{subsec:MatSep}
Material discrimination is the art of identifying the type of material at each pixel in the image. There is some crossover with Image Understanding, but we include it as an Image Preprocessing method, since Image Understanding methods using features derived from the material information might be helpful in improving performance. This has been the case in multi-view X-ray baggage, where material information is more complete~\cite{Bastan2013}.

The interactions of X-rays with a material varies depending on the type of material and the type of radiation. By studying the types of interactions occurring it is possible to identify the type of material by some characteristic such as its effective atomic number. To do this, it is required that multiple energy measurements are made on the material either by illuminating it with multiple radiation sources~\cite{Ogorodnikov2002c}, and/or by using a continuous spectrum of radiation energies and a detector that can resolve the difference in the energy spectrum after interaction~\cite{Gil2011a}. 

Often the high-throughput requirements of commercial systems, mean that they are limited to two energies, a single or few views, and that image noise is sufficient to make it impossible to discriminate between individual atomic elements~\cite{Fu2010b}. Instead researchers attempt to discriminate between groups of materials such as organics, light metals and heavy metals~\cite{Ogorodnikov2002c,Ogorodnikov2002a}. Alternatively some researchers attempt to just identify high-$Z$ materials~\cite{Fu2010,Fu2010b,Fu2010c,Chen2007a} as they can indicate the smuggling of radioactive materials or their shielding. Even in these simple cases, researchers have found it difficult to accurately discriminate materials from raw measurements on a pixel-wise basis, finding that it is necessary to incorporate spatial information into discrimination~\cite{Ogorodnikov2002c}. Thus, researchers have applied a number of image segmentation approaches to aid with discrimination.

The majority of the cargo material discrimination literature uses Dual-energy X-ray systems and are based on $\alpha$-curve~\cite{Novikov1999,Li2016}, $R$-curve~\cite{Ogorodnikov2002c}, or $H$-$L$ curve~\cite{Zhang2005} methods. There has been little influence in cargo work from the baggage or medical domains, due to the much higher energy regime (Fig.~\ref{fig:attenuation}). For example, the seminal work in CT by Alvarez and Macovski~\cite{Alvarez1976}, which expands the attenuation coefficient as a set of intuitive basis functions. This works in the CT energy regime where the photoelectric interaction, which depends strongly on atomic number, is dominant. But it is subservient to pair production and scatter in the cargo energy regime.

The $\alpha$-curve~\cite{Novikov1999,Li2016}, $R$-curve~\cite{Ogorodnikov2002c}, and $H$-$L$ curve~\cite{Zhang2005} methods attempt to estimate the effective atomic number ($Z$) grouping (i.e. organic, light metals, heavy metals) by combining high and low energy transparencies to form a value that can be mapped to effective $Z$ grouping using a lookup table. Authors tend to define the transparency $T$ by normalising the image by the total number of photons (integrated over the range of energies $E$) emitted by the source and the detector sensitivity $D(E)$.

The $R$-curve method is motivated by taking a transparency captured at energy $E_1$ and a second at a different energy $E_2$, and taking the ratio of their logs
\begin{equation}
R(E_1,E_2,Z_0) = \frac{\log(T_1)}{\log(T_2)} = \frac{\mu(E_1,Z_0)}{\mu(E_2,Z_0)}\frac{D(E_1)}{D(E_2)}.
\end{equation}
For the monochromatic and single material case, the $R$-ratio is unique to the material atomic number $Z_0$ and so materials can be discriminated, at least in theory. This method is well-suited to $\gamma$-ray imaging where the photons are emitted with quantised energies. However, in cargo X-ray, the X-ray source is not monochromatic and has a continuous Bremsstrahlung distribution. In this case $R$ varies as a function of the material mass thickness. Nevertheless, one can attempt to recover the effective atomic number grouping at a pixel by experimentally measuring the $R$-ratio as a function of mass thickness to create a lookup table. There are difficulties at low mass thickness where the $R$-ratio versus mass thickness curves for different materials overlap. 

The $\alpha$-curve method computes the quantities
\begin{eqnarray}
\alpha_1 &=& -\log{T_1},\\
\alpha_1-\alpha_2 &=& -\log{T_1}+\log{T_2}.
\end{eqnarray}
Again a lookup table is determined through experimentation.

Finally, the $H$-$L$ curve method simply creates a lookup table using the  high (H) and low (L) energy images $I_1$ and $I_2$.

\begin{figure}[htbp]
\centering
\includegraphics[width=1\textwidth]{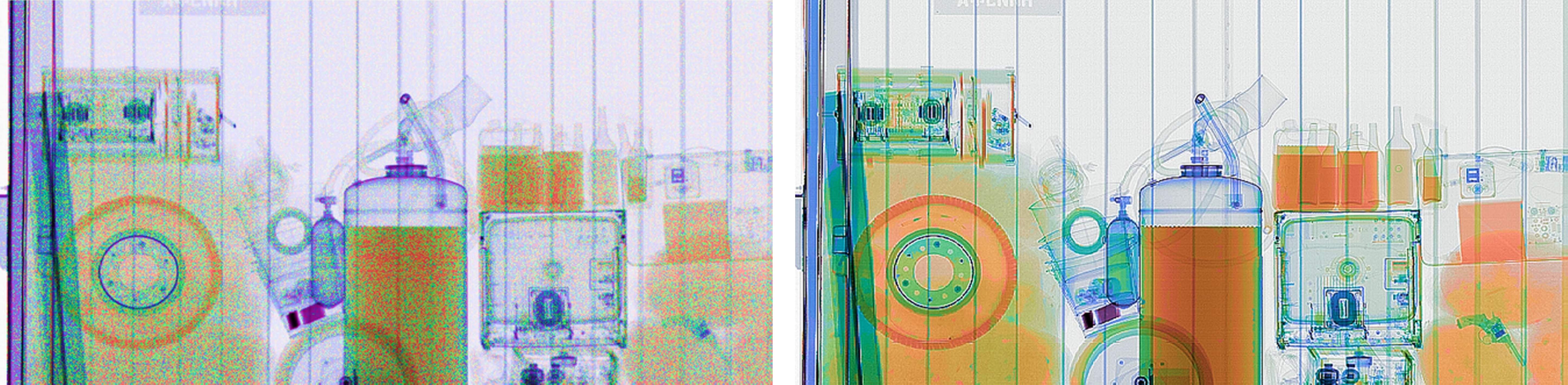}
\caption{Example of material discrimination on individual pixels (left) and on spatial regions (right). Source: Ogorodnikov and Petrunin~\cite{Ogorodnikov2002c}.}
\label{fig:OgorMatSep}
\end{figure}

The seminal work for dual-energy material discrimination for cargo was by Ogorodnikov and Petrunin~\cite{Ogorodnikov2002c,Ogorodnikov2002a}. The authors introduce the $R$-curve method and attempt to classify materials into four groups: organics (hydrocarbon, $Z{\sim}5.3$);  organics/inorganics (aluminium, $Z{\sim}13$); inorganics (iron, $Z{\sim}26$); and heavy substances (lead, $Z{\sim}82$). They use a prototype inspection system, with a $4/8\units{MeV}$ cut-off Bremsstrahlung beam and a lead beam filter. They identify that the $R$-ratio crossover of iron and lead can be translocated by use of the filter, thus allowing improved discrimination for small mass thickness~\cite{Ogorodnikov2002c}. The authors first study the error when discriminating iron from hydrocarbon as function of mass thickness, and find discrimination is optimal at 40-60\units{g/cm$^2$}. They reason that discrimination error increases for lower mass thickness because there is not sufficient contrast between low and high energy images, and for larger mass thickness due to decreasing signal-to-noise. The authors note that, when discriminating between all four groups, material recognition is unreliable, in particular the water-aluminium discrimination error reaches $40\%$ at the optimal mass thickness. To remedy this, they incorporate spatial information using a modified Leader clustering algorithm. The modification ensures that spatially disjointed pixels do not belong to the same cluster. All pixels within a given cluster are labeled as a single material determined by taking the centre of the cluster and comparing to the $R$-ratio lookup table. To avoid over-segmentation, the authors iteratively merge small clusters with larger neighbouring clusters. They use Student's $t$-test to determine which clusters to merge. Coloured material discrimination images with and without incorporation of the spatial information are shown in Fig.~\ref{fig:OgorMatSep}. Qualitatively, it is evident that the use of spatial information greatly improves image quality.

A few years later, Zhang \emph{et al.}~\cite{Zhang2005} introduced the $H$-$L$ curve method. They introduce a \emph{material intrinsic difference} measure, defined as
\begin{equation}
\text{\emph{diff}} = \frac{H_2-H_1}{(H_2+H_1)/2}\times 100\%,
\end{equation}
where $H_1$ and $H_2$ are the high energy reading for two different materials. This is similar to a measure introduced by Ogorodnikov and Petrunin~\cite{Ogorodnikov2002c,Ogorodnikov2002a} but for the $R$-ratio. They use $\text{\emph{diff}}$ for evaluating the material discrimination abilities of a dual-energy scanning system. The authors argue that if image noise is below $\text{\emph{diff}}$, then materials can be accurately discriminated. They give a table of results showing the measured $\text{\emph{diff}}$ for different adjacent-$Z$ materials and find $\text{\emph{diff}}$ to be a decreasing function of $L$. The authors do not show any evidence of applying the $H$-$L$ curve method to whole images.

Since these initial works, other researchers have largely focused on high-$Z$ detection, claiming that multi-group material discrimination is infeasible for commercial systems. For example, Fu \emph{et al.}~\cite{Fu2010b} claim that identifying the effective $Z$ of the scanned objects is not practical because it requires high precision measurements and the noise in commercial systems is too large. Most have focused on the detection and segmentation of suspicious or high-$Z$ materials.

Fu \emph{et al.}~\cite{Fu2009a} attempt to segment suspicious, shielded objects. They introduce a \emph{hybrid clustering} approach which does not require a prior on the number of clusters or the size of clusters, but a prior on the \emph{step level}, which determines the number of quantisation levels in the clustered image given the maximum image value. Hybrid clustering performs clustering followed by region growing. For clustering, each pixel is first compared to the mean of its neighbourhood, if the pixel is close to the mean then its value is assigned as the quantisation of that mean. If it is not close, then they split the neighbourhood into quadrants, compute the means, and set the pixel value to the nearest quadrant's quantised mean. They claim that this is faster than recursive K-means clustering and the Leader clustering used by Ogorodnikov and Petrunin~\cite{Ogorodnikov2002c,Ogorodnikov2002a}. It is beneficial because it creates continuous and not disjoint clusters. After clustering they do region merging, using the highest intensity region as the seed. To segment shielded objects, the authors iterate through the different quantisation levels, binarise the image by quantisation, and then attempting to region fill based on gradients. If the intensity of a filled region is greater than the surrounding, then it is regarded as a shielded object. The method is tested on a cargo image with various amount of shielded lead and tin. No quantitative measure of the performance is given, but the method appears to work well on the single test image presented.

In a separate paper, Fu \emph{et al.}~\cite{Fu2010c} attempt to improve detection and reduce false alarms for high-$Z$ detection. They apply their hybrid clustering described in~\cite{Fu2009a}. After identifying regions that are shielded by low-Z materials, they attempt to separate the shielded object from the background by subtracting the \emph{shielding} attenuation from the \emph{shielded} attenuation. They claim that leads to better high-$Z$ detection. In another paper~\cite{Fu2010b}, they identify two sources of error, namely the \emph{edge effect} at object edges due to scatter, misalignment, digitization, and Poisson noise. They propose use of a wavelet shrinkage denoising approach, which reduces false negatives and false positives, but no quantitative measure of performance is determined. The authors state that similar results can be achieved by use of a Weiner filter, but that it needs to be combined with morphological filtering.

Chen \emph{et al.}~\cite{Chen2007a} also focus on detecting high-$Z$ material. They use a 6/9\units{MeV} commercial system. No substantial details of the methods are given, although they state that the high-$Z$ signature is generated using ``dual-energy information processing, machine vision and topology analysis, and background object striping''~\cite{Chen2007a}. They show an example of lead detection against a piece-wise varying background density, but no quantitative measure of performance is given.

In a recent paper, Ogorodnikov \emph{et al.}~\cite{Ogorodnikov} refer to their original work~\cite{Ogorodnikov2002c} and echo the sentiments of other researchers; that their previous approach to material separation is labile, instable and not repeatable in practical implementation. In this paper, although their algorithmic methods are not detailed in full, the authors attempt 3-group (organics, mineral/light metals, metals) material discrimination but this time with a 3.5/6\units{MeV} Bremsstrahlung beam. Additionally, they attempt to calculate the mass of the object under inspection. They claim a mass preciseness of ${<}10\%$ and effective atomic number preciseness of ${\pm}1$ in the optimal mass thickness range.

Recently, Li \emph{et al.}~\cite{Li2016} have proposed a solution to improve material recognition when two materials overlap in an image. The method requires prior information about one of the overlapping materials, which the authors argue is available in a practical setting from the shipping manifest, or if trying to separate container and contents an assumption can be made about the container material. Their algorithm firstly performs a pre-classification based on the $\alpha$-curve method, they then determine if a region is more likely composed of a pure material or two overlapping materials. If composed of two materials, the next step decomposes the material into the two overlapping contributions. The final step is to perform recognition on the materials. To decompose overlapping materials, the authors use a method originating from Dual-Energy X-ray Absorptiometry (DEXA) which is used for measuring bone mineral density and soft-tissue composition of human bodies. The method uses second-order conic surface equations to approximate the polychromatic transparencies of the high and low energy images. They fit the conic surface parameters using least-squares. The authors test the algorithm on synthesised data and real data captured in a lab experiment, and achieve good qualitative results.

Other researchers have investigated the possibility of material discrimination on systems that are not dual-energy, using simulations. For example, Gil \emph{et al.}~\cite{Gil2011} use Monte Carlo simulation but to investigate the possibility of \emph{single-shot material discrimination}. The single-shot method assumes that the detectors can measure the energy spectrum of the beam and can split it into a low and high energy component to determine the $R$-ratio. The authors simulate a Bremsstrahlung beam with 9\units{MeV} cut-off, the low-high division is chosen as 4\units{MeV}. They compare the one-shot $R$-ratio to a 4/9\units{MeV} dual-energy simulation. Comparing the $R$-ratio for silver and tissue-equivalent plastic, it appears that the one-shot method has a greater discriminative effect, whilst potentially having the benefits of lower X-ray dose and faster scan time. However, it is unclear how this method would work in practice since the derivation of the one-shot $R$-ratio requires splitting of the energy spectrum before interactions with the scene. Furthermore, it is unclear whether the system model results in a realistic level of noise when compared to a commercial system.

Fantidis \emph{et al.}~\cite{Fantidis} investigate potential mixed $\gamma$- and X-ray system architectures, and their ability to discriminate materials, through Monte Carlo simulation. They simulate three $\gamma$ sources (\nuclearSym{60}{Co}, \nuclearSym{137}{Cs}, and \nuclearSym{88}{Y}), and a 4/9\units{MeV} dual-energy Bremsstrahlung beam. They test material discrimination performance on 165 materials and using different dual, triple and quadruple combinations of the sources. They assess the potential performance of the system using the number of $R$-overlaps between  different materials. They claim that the optimal selection of sources are 4\units{MeV} Bremsstrahlung and \nuclearSym{137}{Cs} for dual, and 4/9\units{MeV} Bremsstrahlung and \nuclearSym{137}{Cs} for triple. The optimal quadruple source system, although not specified, only offers a slight improvement over the optimal triple source system. There is no evidence that the authors attempt to model system noise and the effects on discrimination, other authors have found that $R$-values alone are not a good indicator of performance due to noise in the $R$-estimates when interrogating materials with small or large mass thickness~\cite{Ogorodnikov2002c}.

\subsection{Segmentation for CT baggage}
In CT baggage, there have been several proposals for single- and dual- energy segmentation, with some based on ML which we review here. The algorithms are designed for segmenting 3D volumes but aspects of the approaches may be transferable to 2D cargo. In CT baggage segmentation, algorithms must cope with a variable and unknown number of baggage items, with large variability in their shapes, types and sizes~\cite{grady2012automatic}. This is in contrast to the medical domain, where segmentation tasks are prespecified, for example a segmentation of a particular organ~\cite{grady2012automatic}. Therefore, baggage researchers have looked to design unsupervised algorithms that make no assumptions on the number of objects or on their composition.

The approach taken by Grady \emph{et al.}\cite{grady2012automatic} for single-energy CT, first identifies object voxels, then identifies candidate object splits using the Isoperimetric Distance Tree (IDT) method~\cite{grady2006fast}, and finally evaluates good splits according to a novel Automatic QUality Assessment (AQUA) metric learnt from a large training set. The initial coarse segmentation uses a Mumford-Shah based method~\cite{grady2009piecewise} applied to a preprocessed (denoised and artefact reduced) CT image. The AQUA method is based on 42-dimensional descriptor from the prior literature on object segmentation, and includes features based on geometry, intensity, gradients and ratios of those features. To learn the AQUA model, the authors first use Principal Component Analysis (PCA) to reduce to reduce dimensionality. They fit a Gaussian Mixture Model (GMM) over the PCA coefficients of all the segments in the training set using Expectation-Maximisation (EM). Aqua is used both to select best candidate splits, and to select the best segmentation over three different parameter settings.

Mouton \emph{et al.}\cite{mouton2015materials} introduce a material-based segmentation for low resolution Dual-Energy CT (DECT) images representative of the aviation security environment. After preprocessing to reduce metal artefacts, the authors first perform a coarse segmentation based on the Dual-Energy Index (DEI) and connected component analysis. The DEI combines the high and low energy linear attenuation coefficients at each voxel to give a crude estimate of the material characteristics. The authors use a Random Forest (RF) model to guide the segmentation process by assessing the quality of individual object segments and the entire segmentation. For individual object segments, the trained RF model uses the same 42-dimensional descriptor used by Grady \emph{et al.}\cite{grady2012automatic}. The authors claim that using the RF approach outperforms AQUA in their aviation setting. The quality of full segmentations is assessed using the RF score of constituent objects weighted by the error in the number of segmented objects. The authors demonstrate that their approach outperforms three state-of-the-art segmentation techniques, including: IDT~\cite{grady2006fast}; Symmetric Region Growing (SymRG)~\cite{wan2003symmetric}; and 3D Flood-Fill region growing (FloodFill)~\cite{wiley2012automatic}.

In cargo, material-based segmentation is much more challenging due to the limitation of 2D, the overlapping of materials and objects, and the inability to reconstruct linear attenuation coefficients that encode material information. However, the $\alpha$-curve~\cite{Novikov1999,Li2016}, $R$-curve~\cite{Ogorodnikov2002c}, and $H$-$L$ curve~\cite{Zhang2005} methods can provide crude (more so than DEI) material information that could potentially be used to initiate coarse segmentations. Similar methods to AQUA~\cite{grady2012automatic} and the RF model of Mouton \emph{et al.}~\cite{mouton2015materials} could be used to identify object splits and overall segmentation quality. However, it is likely that extra metrics are required to take care of overlapping objects without \emph{a priori} information on the number of objects overlapping or their characteristics such as thickness and material. Methods have been proposed in multi-view baggage for layer separation that may be applicable to multi-view cargo~\cite{heitz2010object}. To date, we are not aware of any proposals for cargo, or indeed single-view baggage, that can convincingly address these issues.

\subsection{Discussion on Image Preprocessing}
\label{subsec:discussionImagepreprocessing}
Of the topics identified in Image Preprocessing, by far the most work has been on material discrimination. The methods are largely derived from physics, and as far as we know, no ML techniques (similar to baggage Refs.~\cite{mouton2015materials,grady2012automatic}) have been applied to the subject due to the difficulty of obtaining sufficient data with accurate labeling. Additionally, since all authors tend to use different datasets from different commercial partners, or independent lab experiments, it is difficult to compare the performance between different contributions. Furthermore, most authors choose to evaluate performance qualitatively rather than quantitatively, and often using only a single image. We feel that researchers need to better quantify per-pixel classification performance so that different methods can be more easily compared. Moreover, we believe that the field would benefit from an open dataset available for researchers.

There has been three main methods introduced for initial pixel classification; the $R$, $\alpha$, and $H$-$L$ curve methods. It is not immediately obvious which should perform best, or if in fact they all perform equally. This is because researchers have not yet performed a comparison of the different methods on the same dataset. Such a study is a future avenue for research in the area. Particularly, when a new method is introduced, it should be compared to the methods already existing in the literature.

For image manipulation and image quality improvement, there is a need to evaluate, compare and understand different techniques in terms of their effect on the performance of machine and human Image Understanding. Such work has been attempted in the baggage domain~\cite{Mouton2013}. For TIP, some work in cargo has been done as a preprocessing step for training automated Image Understanding algorithms, and TIP methods have only just been put through basic experimental validation by Rogers \emph{et al.}~\cite{rogers2016threat}. The effects of training ML algorithms on synthesised threat images are yet to be fully understood. 

Although no work has been done on verifying that ML algorithms trained on TIP-augmented cargo data actually boosts performance, there has been evidence from other fields for many tasks~\cite{Chen2016,Gupta2016}. There are also several problems that still remain. For example, it is difficult to generate out-of-plane rotations, so augmentation is usually limited to in-plane rotations of the staged threat items. Potential remedies include either developing a framework for collecting the optimal number of threat poses to make accurate interpolation of intermediate out-of-plane rotations, or generating realistic threat radiography from realistic 3D CAD models of threats~\cite{White2008a,gong2016rapid}. The requirements for a solution are that out-of-plane rotations are accurate, realistic, and can be computed efficiently or on-the-fly. Whilst the interpolation approach would be fast it may be difficult to obtain good accuracy without capturing a large number of projections due to the complicated fan-beam geometry. Conversely, the CAD approach would enable more accurate computation of out-of-plane poses, but it is unclear how realistic the generated threat image would be and how fast it can be computed if accurate photon transport models are required.

\section{Image Understanding}
\label{sec:imageUnderstanding}
Automated Image Understanding tasks in cargo are currently split into the themes of Automated Contents Verification (ACV) and Automated Threat Detection (ATD). We give an overview of the most pertinent works in the literature in Table~\ref{tab:literatureReview}.

\begin{table}
\footnotesize
\begin{tabular}{ L{0.2\textwidth}L{0.1\textwidth}L{0.3\textwidth}L{0.3\textwidth}}
\hline
	\bf{Study} & \bf{Task} & \bf{Methods} & \bf{Notes} \\ \hline
	Chalmers~\cite{Chalmers2007a,Chalmers2007} & ECV & Intensity hist. metrics (min, max, mean, Std.); compare with historical database example. & No QE given \\ 
	& & & \\
	Orphan et al.~\cite{Orphan2005} & ECV & Segment floor, walls \& roof; rule-based object detection & $97.2\%$ Acc.; $0.4\%$ FPR \\ 
	& & & \\
	Rogers \emph{et al.}~\cite{Rogers2015} & ECV & $96\units{Px}$ windows; image moments, oBIF hist., window coordinates; RF classification; trained on synthesised non-empties. & $99.3\%$ DR and $0.7\%$ FPR on stream-of-commerce; $90\%$ DR and $0.17\%$ FPR on ${\sim}1.5\units{kg}$ cocaine; $90\%$ DR and $0.51\%$ FPR on ${\sim}1\units{L}$ water\\ 
	& & & \\
	Andrews \emph{et al.}~\cite{Andrews2016} & anomaly detection \& ECV & Down-sampled images; sparse auto-encoder; hidden layer features; RBF-SVM & $99.2\%$ Acc.; For features, hidden representation $>$ normalised squared residual \\ 
	& & & \\
	Zhang \emph{et al.}~\cite{Zhang2014} & MV & Leung-Malik filter codebook; SIFT; dense sampling; edge sampling & visual codebook method$>$SIFT. Edge sampling$>$dense sampling.  \\ 
	& & & \\
	Tuszynski \emph{et al.}~\cite{Tuszynski2013} & MV & Median intensity hist.; average absolute deviation; weighted city block distance. & 48\% Acc. and 5\% FPR \\ 
	& & & \\
	Jaccard \emph{et al.}~\cite{Jaccard2014} & ATD; cars & oBIF hist.; intensity hist.; log-intensity hist.; RF classification & 100\% DR and 1.23\% FPR with oBIF. oBIFs$>$log-intensity hist. $>$ intensity hist. \\ 
	& & & \\
	Zheng \emph{et al.}~\cite{Zheng2013a} & ATD & Correlation coefficient; threshold & No QE given, detected anomalies may not correspond to presence of a threat. \\
	& & & \\
	Jaccard \emph{et al.}~\cite{Jaccard2015} & ATD & 9-layer CNN; 19-layer CNN; oBIFs + RF; augmented dataset & 90\% DR 0.8\% FPR. CNNs${>>}$RF+oBIFs \\ \hline\hline
\end{tabular}
\caption{A summary of the literature on automated cargo Image Understanding research, in terms of the task, methods used. Abbreviations: histogram (hist.); Quantitative Evaluation (QE); Accuracy (Acc.); False Posititve Rate (FPR); Detection Rate (DR); Convolutional Neural Networks (CNNs); Scale-Invariant Feature Transform (SIFT); oriented Basic Image Features (oBIFs); Random Forest (RF); Radial Basis Function Support Vector Machine (RBF-SVM). The $>$ symbol denotes `peforms better than', and ${>>}$ denotes `performs much better than'.}
\label{tab:literatureReview}
\end{table}

\subsection{Automated Contents Verification}
\label{subsec:ACV} 
\begin{figure}[htbp]
\centering
\includegraphics[width=\textwidth]{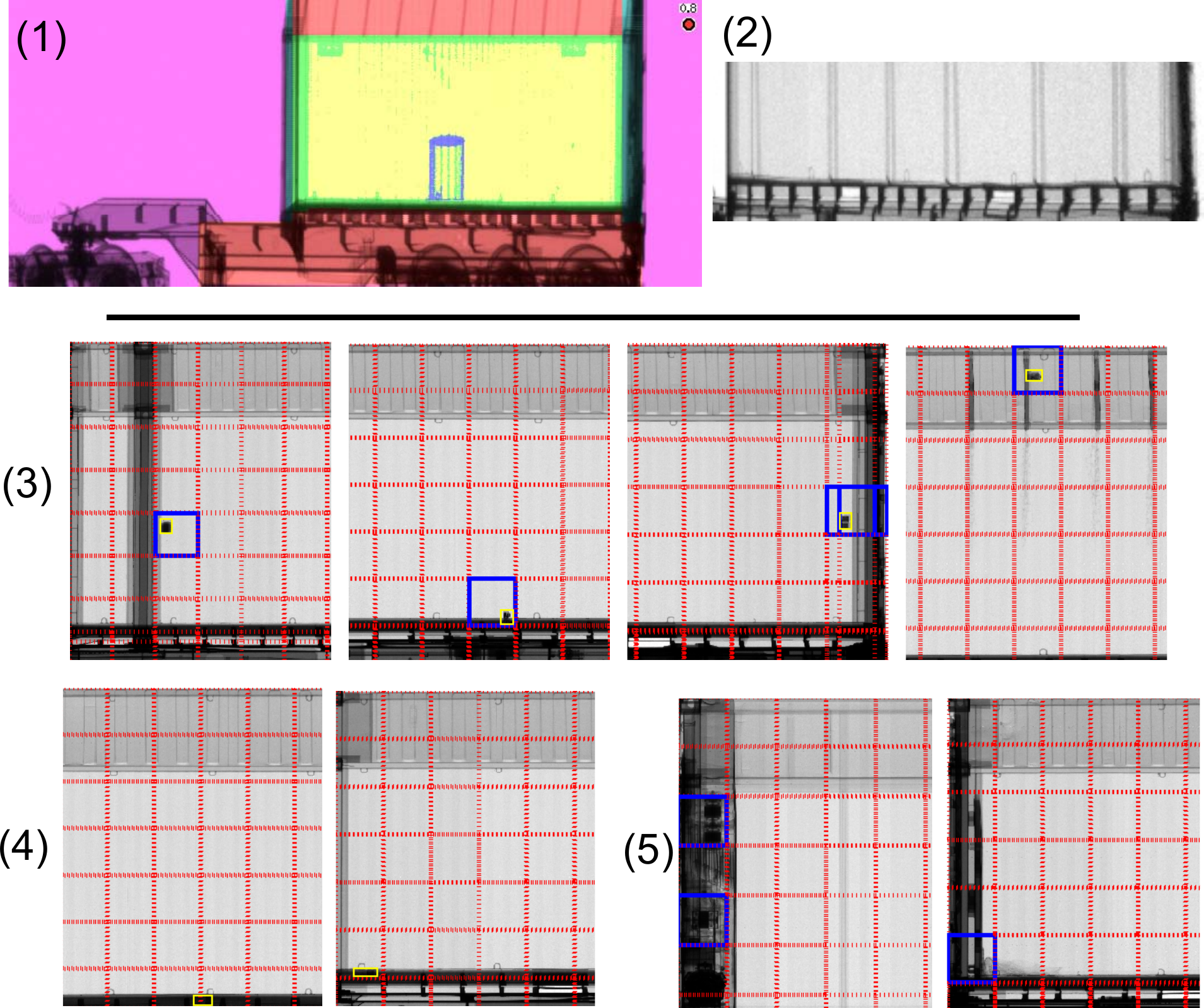}
\caption{Example Empty Cargo Verificaiton (ECV) results of a rule-based algorithm (1 \& 2) and a Machine Learning (ML) based algorithm (1-3). The colouring in image (1) shows empty space (purple), roof and chassis (red), cargo walls (green) and an object (blue). Image (2) shows a false positive caused by the vertical spars. The ML-based algorithm detections are shown in (3), false negatives in (4), and false positives in (5). The red dashed boxes indicate windows classified as empty, and the solid blue boxes show windows classified as containing a load. Sources: Orphan \emph{et al.}~\cite{Orphan2005a} and Rogers \emph{et al.}~\cite{Rogers2015}.}
\label{fig:emptyExamples}
\end{figure}

ACV checks whether the cargo contents match those stated on the shipment manifest. This can range from Empty Cargo Verification (ECV) to full Manifest Verification (MV). ECV  can be useful for increasing throughput, since declared-as-empty cargoes (20\% of all cargo) can be sent through a separate automated inspection lane. ECV examples are given in Fig.~\ref{fig:emptyExamples}. Containers may be falsely declared as empty in shipping fraud, or may be exploited in rip-on/rip-off smuggling operations. False declared-as-empty cargo containers can pose safety hazards during container stacking at ports due to the unexpected additional weight. MV compares the X-ray image to the Harmonized System (HS) codes declared on the manifest. Each HS code defines a different broad category of cargo type, for example, live animals, animal products or vegetable products.

The first work on ECV was Chalmers \emph{et al.}~\cite{Chalmers2007a,Chalmers2007}, who use ``readily available'' algorithms to segment the container region and compute metrics that are then compared with empty containers of the same size. No specific details are given on the algorithms or their performance, but we interpret Ref.~\cite{Chalmers2007} as follows. The container is classified by generating an intensity histogram of the segmented cargo region and comparing to histograms from historical empty images. The comparison is made using metrics such as minimum, maximum, mean, and standard deviation. Another method is briefly described by Orphan et al.~\cite{Orphan2005}, which segments the image (e.g floor, walls, and roof) and then applies an unspecified rule-based object detection algorithm. The authors report $97.2\%$ accuracy (with $0.4\%$ false negatives) when classifying SoC images as empty or non-empty. 

More recently, Rogers \emph{et al.}~\cite{Rogers2015}, have attempted ECV by detecting loads within cargo containers. They claim that ECV is difficult due to the container parts that locally appear similar to small loads, and due to variation in container types (e.g. refrigerated units, bulk units, 20 ft or 40 ft General Purpose). The task is further complicated by container damage and detritus, which the algorithm must learn to ignore. Their method splits the image into a grid of small $96{\times}96\units{pixels}$ windows. Then for each window they compute image moments and oriented Basic Image Features (oBIFs) at a range of scales. They feed the features, along with the window spatial coordinates into a Random Forest (RF). The authors claim that the spatial coordinates allow the RF to implicitly learn the range of possible empty container appearances at different locations. The classification decision for the image is determined by taking the maximum score of the windows composing the image and comparing it to a tunable threshold. The authors generate synthetic examples (TIP) of non-empty containers in order to train the algorithm, this allows training on more difficult examples than those found in the SoC. The algorithm is tested on both real SoC data and difficult synthetic examples. On the SoC data, it is able to detect 99.3\% of non-empty containers while raising 0.7\% false alarms on truly empty containers. On difficult examples they are able to achieve 90\% detection for loads similar to 1.5\units{kg} of cocaine or 1\units{L} of water, while raising 1-in-605 or 1-in-197 false alarms, respectively.

Andrews \emph{et al.}~\cite{Andrews2016} have recently used ECV as a test problem for anomaly detection using auto-encoders. They use cargo X-ray images of empty and non-empty containers down-sampled to $32{\times}9\units{pixels}$, and in one test take the empty containers (tight appearance) as the normal class and in another they take the non-empty containers (diverse appearance) as the normal class. The authors derive a number of features from the hidden layer of a trained sparse auto-encoder, including: the hidden representation, the scalar residual magnitude; the signed residual (with and without normalisation by the root-mean-squared residual); the absolute residual; and the squared residual (with and without normalisation by the mean-squared residual). The features are classified using a one-class Radial Basis Function Support Vector Machine (RBF-SVM). When considering non-empty containers as the normal class, they find that the RBF-SVM achieves best classification accuracy (92.99\%) when fed the hidden representation as a feature. When considering empty containers as the normal class the best accuracy (99.2\%) is achieved when the normalized squared residual is used as the feature. 

There have been two published attempts at MV~\cite{Zhang2014,Tuszynski2013}. MV is a multi-class classification task, where cargo containers are classified according to HS code. Tuszynski \emph{et al.}~\cite{Tuszynski2013} compute the median image grey-level histogram and average absolute deviation to form a model for each HS code. They then use a weighted city block distance to compare a given example to each HS code model. This approach yields an overall accuracy of 48\% given a false positive rate of 5\%. This result is improved slightly by Zhang \emph{et al.}~\cite{Zhang2014}, who use a Leung-Malik filter bank to construct a visual codebook as a texture descriptor. They determine that this outperforms Scale-Invariant Feature Transform (SIFT) when classifying cargo images according to their HS code. Note that the authors ignore ``non-classical'' examples, which they define as those containers that are less than half filled with cargo. We feel that for real-life deployable system, such examples should be included since an adversary could purposefully choose to only half fill a container when smuggling or to avoid duties.

\subsection{Automated Threat Detection}
\label{subsec:ATD}
Currently, there are few publications on cargo ATD, much more work has been done for baggage screening. The first such paper was on detecting cars that may be stolen or undeclared to avoid duties. Jaccard et al.~\cite{Jaccard2014} use oBIF histograms computed at a range of scales and a RF classifier. They oversample car windows to boost the number of car examples in the training set. Using a Leave-One-Out-Cross-Validation (LOOACV) scheme they determine a detection rate of $100\%$ of car-containing containers while raising $<1\%$ false positives on SoC non-car containers. The authors also investigate other features such as intensity histograms, log-intensity histograms, and Basic Image Features (BIFs), but found these inferior to using oBIFs. In a later paper~\cite{Jaccard2016}, the authors were able to improve performance to 100\% detection rate for a false alarm of 0.41\%, by including more oBIF scales.

Zheng and Elmaghraby~\cite{Zheng2013a} propose a method for ATD in vehicles by detecting anomalous regions within images. They use backscatter images (top view and two side views) and a transmission image (side view) captured from an AS{\&}E OmniView\textregistered\ Gantry. They perform a window-wise correlation analysis comparing a fresh image of the vehicle to a historical image of the same vehicle stored in database. Images are split into 64 rectangular $4{\times}16\units{pixel}$ windows, and the correlation between the same windows in the fresh and historical image is computed. Correlation is also computed for fresh image windows with windows from different locations in the historical image, to account for goods that may have moved around inside the vehicle. In total, they compute a $64{\times}64$ matrix of window correlation values. A given window is classified as anomalous if the maximum of the corresponding matrix row is below a threshold. No quantitative evaluation of the performance is given. A criticism of this proposed method is that an anomalous region will very rarely indicate an actual threat and so the false positive rate is likely to be extremely high.

\begin{figure}[htbp]
\centering
\includegraphics[width=0.7\textwidth]{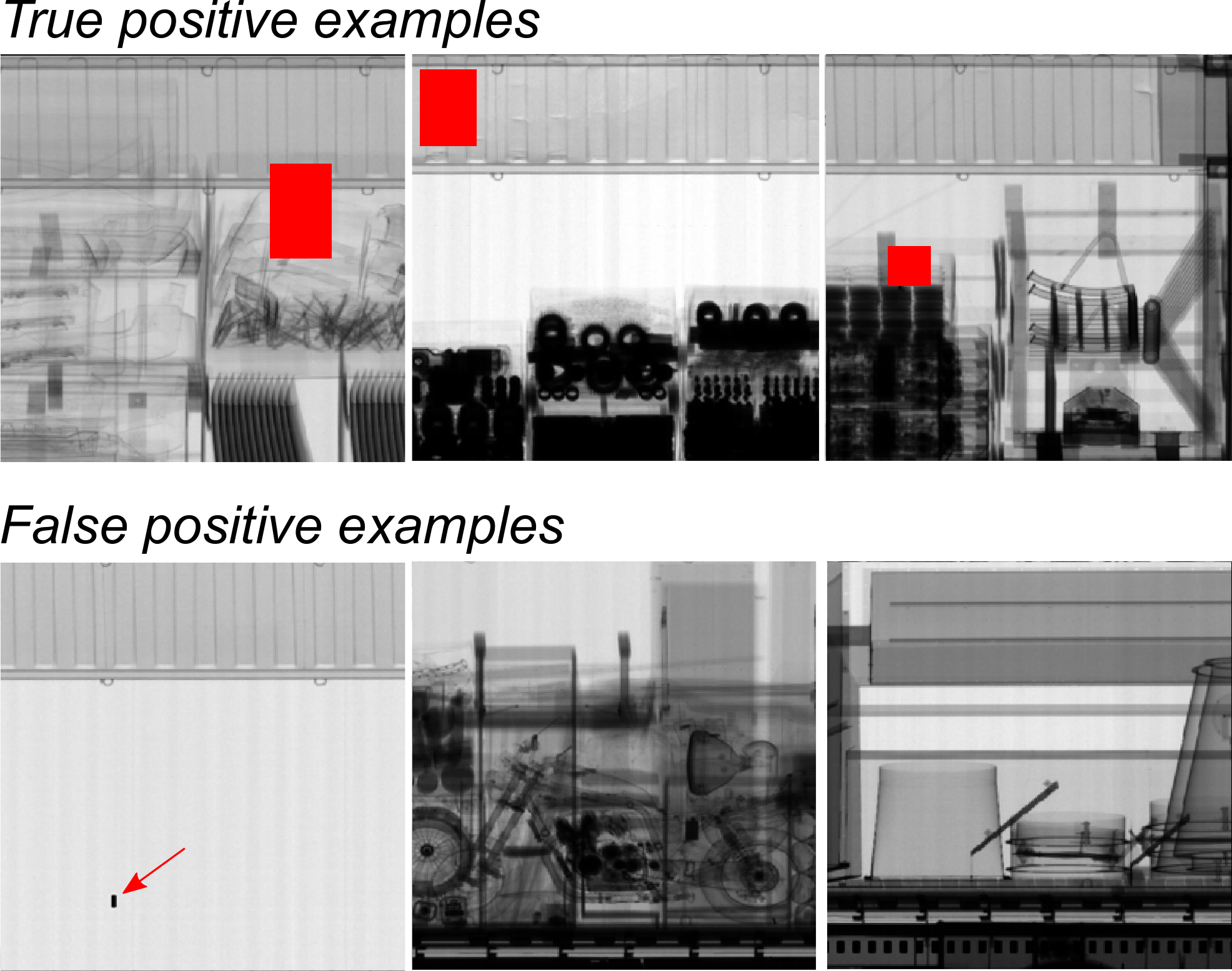}
\caption{Examples of Small Metallic Threat (SMT) detections and false positives using a trained-from-scratch Convolutional Neural Network (CNN). The original authors have censored the SMTs with red boxes. The SMTs are long and thin, and so the censoring boxes indicate approximate length and orientation of objects. Source: Jaccard \emph{et al.}~\cite{Jaccard2015}.}
\label{fig:SMTDetection}
\end{figure}

Jaccard \emph{et al.}~\cite{Jaccard2015} attempt to detect threats that are ``akin to small metallic objects (e.g. drill)''; the exact nature of the threats are censored to prevent keyword searching. The method uses CNNs trained-from-scratch on an augmented dataset, with real threat images projected into images from the SoC (TIP). The authors found that a 9-layer shallow network architecture (Krizhevsky \emph{et al.}~\cite{Krizhevsky2012}) and a very deep 19-layer architecture (Simonyan and Zisserman~\cite{Simonyan2014}) lended themselves well to the task. The shallow network uses convolutional layers with large receptive fields, and each followed by a max pooling layer. Whereas the very deep network uses convolutional layers, with small receptive fields, and stacked in twos or threes between each max pooling layer. In both cases the classification decision from the fully connected output layer is made using the softmax function. The authors compare the CNNs to a oBIF+RF method similar to that previously used to detect cars~\cite{Jaccard2014}. Both the shallow and very deep network provided a huge boost in performance over oBIF+RF, with the very deep network performing slightly better than the shallow network. The authors report a false alarm rate of $0.8\%$ given $90\%$ detection. Examples of SMT results for the CNN approach are given in Fig.~\ref{fig:SMTDetection}.

\begin{figure}[htbp]
\centering
\includegraphics[width=\textwidth]{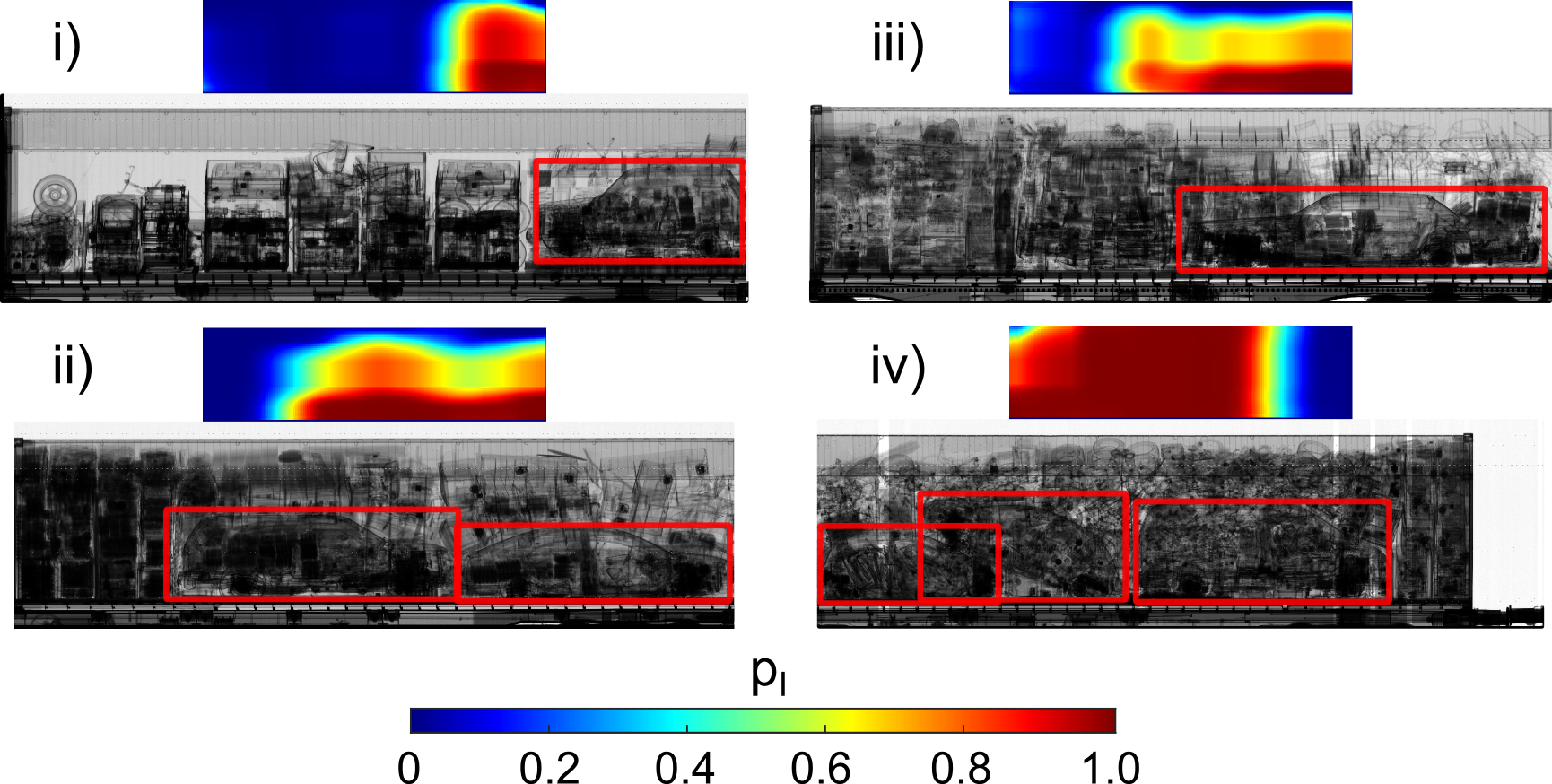}
\caption{Examples of obscured car detection for using a trained-from-scratch Convolutional Neural Network (CNN). The heatmaps show the average classifier confidence $p_{I}$ for a window sliding across the image. Source: Jaccard \emph{et al.}~\cite{JaccardCars16}.}
\label{fig:carDetection}
\end{figure}

Most recently, Jaccard~\emph{et al.}~\cite{JaccardCars16} have revisited their car detection work~\cite{Jaccard2014} and applied a trained-from-scratch very deep 19-layer CNN~\cite{Krizhevsky2012}. The authors again use window oversampling to increase the number of car training examples. A method based on Pyramid Histograms of Visual Words (PHOW) was also assessed. The authors find that the CNN approach yielded 100\% detection and 0.22\% false alarms, and was able to detect even heavily obscured cars. Moreover, the CNN approach yielded 5-fold and $\frac{3}{2}$-fold improvements in false alarm rate over the PHOW-based method and oBIF+RF method used in Ref.~\cite{Jaccard2014}. Examples of car detection results are given in Fig.~\ref{fig:carDetection}.

\subsection{ATD for baggage}
\label{subsubsec:ATDBaggage}
More ATD research has been carried out in baggage, and detailed summaries can be found in the review by Mouton \emph{et al.}~\cite{Mouton2015a}. We give a brief overview of the points important to cargo.

Several different X-ray imaging modalities are used in baggage screening. These range from single-view~\cite{Riffo2015}, to multi-view~\cite{Mery2012NDT,Mery2013,Mery2013a,franzel2012object,Bastan2013}, to full 3D Computed Tomography (CT)~\cite{Flitton2015,Mouton2014,Flitton2013,Flitton2010}. Classification performance typically improves from single view to CT as more information becomes available. The challenge is how to best use this information.

The general consensus amongst the baggage community, is that classification based on X-ray image data is more challenging than visible spectrum data, and that direct application of methods frequently used in natural images (such as SIFT, Rotation Invariant Feature Transform, and Histogram of Oriented Gradients) do not perform well~\cite{Bastan2011}. However the performance can be improved by utilising the characteristics of X-ray baggage images. For example researchers have found that object detection can be improved by augmenting multiple views, using a false colour material image (where pixels are coloured according to the type of material)~\cite{Bastan2015}, or using simple descriptors such as density histogram (DH) or density gradient histogram (DGH)~\cite{Flitton2015,Flitton2013}. 

While it has been widely reported that texture descriptors in baggage scans perform poorly due the lack of texture in X-ray examples~\cite{Schmidt-hackenberg2012,Bastan2015,Bastan2011}, the amount of texture visible in cargo X-ray images \emph{does} differ significantly between images. Medium to low density cargo (such as tyres, and machinery) often contain a lot of texture, while high density cargo (such as barrels of oil) has a more uniform appearance. This is possibly why researchers in cargo have enjoyed more success with texture descriptors such as oBIFs~\cite{Rogers2015,Jaccard2014} or visual codebooks based on a Leung-Malik filter bank~\cite{Zhang2014}.

Franzel \emph{et al.}~\cite{franzel2012object} propose a method of fusing detection results from multiple single views to exploit the extra information from multi-view. They use a voting-based scheme where detection confidence is increased if rays from detection points from single views intersect in 3D. The motivation is to suppress false alarms since they do not coincide in different views, and to reinforce detections that do. The detection confidence on the single view images are determined by sliding a window over the image, computing Histogram of Oriented Gradients (HOG) as features and using a linear SVM. They address in-plane rotations using a non-maximum suppression scheme, since HOG features are not rotation invariant. Moreover, they claim that the multi-view voting fusion scheme handles out-of-plane rotations. They achieve significantly better detection with their multi-view scheme (80\%) over single view (50\%) for 50\% false alarm rate.

Ba{\c{s}}tan \emph{et al.}~\cite{Bastan2013} propose a different multi-view approach. Instead of fusing single-view classifier confidences, they fuse single-view features. The authors experiment with sparse interest point detectors and dense sampling, with SIFT descriptors and its derivative (GLOH, CGLOH and CSIFT), as well as the domain spin image descriptor (SPIN) and two novel variants; ESPIN and CSPIN which incorporate energy information. ESPIN is the concatenation of SPIN descriptors computed on the high and low energy images separately, and CSPIN is the concatenation of SPIN descriptors computed on each channel of the material-coloured image. The authors use a linear Structural SVM (S-SVM) with a branch-and-bound subwindow search framework, which is shown to be more efficient than classical sliding windows. They found both ESPIN and CSPIN performed better than SIFT and SPIN alone, with CSPIN achieving best performance. Like Franzel \emph{et al.}~\cite{franzel2012object}, Ba{\c{s}}tan \emph{et al.}~\cite{Bastan2013} find that their multi-view feature concatenation approach performs better than single view. Moreover, their approach performs significantly better than the approach adopted by Franzel \emph{et al.}~\cite{franzel2012object}.

Multi-view fusion approaches similar to those proposed by Ba{\c{s}}tan\emph{et al.}~\cite{Bastan2013} and Franzel \emph{et al.}~\cite{franzel2012object} might be applicable to multi-view fusion in cargo, however performance is likely to be far worse due to the additional complexity. We feel that a possible approach to multi-view detection, for both baggage and cargo, would be to feed the different views into a CNN as separate channels or separate streams. The CNN can learn to jointly use information from the separate views to make better classifications. For 3D shape recognition, Su \emph{et al.}~\cite{Su_2015_ICCV} have found that CNNs fed with multiple 2D views as inputs performs better than using state-of-the-art 3D shape descriptors. It would be an interesting study for ATD in CT, particularly if better performance can be obtained without having to reconstruct the full 3D baggage image.

Recently, A\c{c}kay \emph{et al.}~\cite{Ackay2016} have applied CNNs to ATD in single-view baggage imagery. They recognise that there is a problem with training CNNs from scratch due to the limited availability of data. Thus they adopt a transfer learning approach by taking a pre-trained CNN, primarily trained for general image classification tasks, and fine-tune it for ATD in X-ray baggage. The pre-trained CNN follows the architecture introduced by Krizhevsky \emph{et al.}~\cite{Krizhevsky2012}, consisting of 5 convolutional layers, 3 fully-connected layers and trained on the ImageNet dataset. The authors re-use the generalised feature extraction and representation in the lower layers of the CNN, whilst fine tuning the upper layers. This achieves 99.26\% detection and 0.74\% false positives, which significantly outperforms prior work in the field. The authors do not comment on the possibility of training a CNN from scratch on data augmented with TIP imagery and realistic variation, such as the work in cargo by Jaccard \emph{et al.}~\cite{Jaccard2015}. Since TIP methods are well-developed for baggage imagery, it would be an interesting comparison to make between a pre-trained and a trained-from-scratch CNN.

\subsection{Discussion on Image Understanding}
\label{subsec:discussionImageUnderstanding}
It has been just over a decade since publications started to emerge on cargo Image Understanding. In initial works, algorithms were typically based on computing simple features (such as maximum image intensity) and applying intuitive hard-coded rules~\cite{Orphan2005}, or by simple comparisons of an image with historical images from a database~\cite{Zheng2013a,Chalmers2007a,Chalmers2007}. Since these initial works, researchers have started to apply ML methods to learn the rules, and even features, from data. Researchers have found that limited access to large, labeled, datasets is still a problem and have started to use Threat Image Projection (TIP) to increase the total amount of training data and the amount of variation within it~\cite{Rogers2015,Jaccard2015,Jaccard2016}. Other researchers, in baggage, have chosen to take CNN models trained for recognition tasks on natural images, and fine-tune them for high performance on X-ray imagery~\cite{Ackay2016}. 

The use of Deep Learning methods, such as CNNs, where feature extraction, representation and classification is learnt simultaneously, shows great promise~\cite{Ackay2016,Jaccard2016}. Such methods have been shown to achieve superhuman performance in a number of visual tasks, including face recognition and image categorisation~\cite{Krizhevsky2012}. It is, therefore, perfectly acceptable to believe that these methods can, and will, outperform humans at visual inspection of X-ray images. The main obstacle to achieving this is the lack of a very large cross-vendor SoC dataset complete with labels, from which a CNN can be trained from scratch and compared to baseline professional human operator performance.

We feel that the main problem with the cargo Image Understanding field is the lack of open datasets for researchers to score and compare methods on. Although, it is unlikely that such datasets will be made available for threat items such as weapons, datasets could be made available which contain benign non-sensitive items. If the dataset was labeled with anonymised manifest information (e.g. HS-codes), we feel it would provoke wider interest in the field, since the X-ray cargo images are a very different problem to natural images. 

There are many avenues for future research in the field, due to its relative infancy. It would be interesting to see how Deep Learning based object categorisation and semantic segmentation methods work on X-ray cargo images. Such methods could find good use as a form of Automated Contents Verification in Assisted Inspection or Selection. In particular, customs agencies store very large collections of cargo images complete with manifest information (labeling), which would be ideal for training CNNs from scratch. However, these datasets are notoriously difficult for researchers to gain access to. Alternatively, transfer learning approaches similar to A\c{c}kay \emph{et al.}~\cite{Ackay2016} could be explored.

Another future challenge, is to develop generalised algorithms that work on images from multiple scanning architectures. So far, algorithms have been developed for a single type of scanner from a single vendor. As far as we know, no researchers have evaluated their algorithms on images from different scanners, and so it is not evident that algorithms would generalise well. Generalisation might be achievable by using transfer learning methods to fine-tune algorithms to specific scanning architectures, or by developing data augmentation techniques that transform images so that they appear as if captured from different, random, scanning architectures. 

\section{Conclusion}
\label{sec:conclusion}
Automated Analysis of cargo X-ray imagery is still a relatively young field. Over the last decade, more attention has been paid to aviation image analysis (such as baggage), since problems are generally more tractable, and because there has been more funding directed towards aviation due to the more perceivable immediate threat from terrorism. Typically, most work in cargo has been kept in-house by industry for commercial and security reasons. However, academics are beginning to form relationships with industry partners, gaining access to large image datasets with which to work.

In comparison to natural images, cargo X-ray images offer an interesting and difficult challenge for researchers, since objects are translucent making occlusions difficult to disentangle, are usually very cluttered and noisy, whilst appearing a skewed in perspective due to the geometry of the X-ray beam. Furthermore, image contents are often more varied than images from the baggage or medical X-ray imaging domains, since a very diverse range of objects are shipped inside containers. We believe that more researchers would become involved in the field if data was easier to get hold of, for example, through the creation of large, labeled, open datasets.

During this review, we have identified several open questions, and avenues for future research, which we now summarise. 

First, there is need for a comparison study of different image preprocessing techniques (i.e. denoising, manipulation and correction), and their effects on the performance of human and algorithmic Image Understanding needs to be understood. It might be that for CNN-based methods, denoising is not essential, however that performance can be improved considerably using some image manipulation. There is a hint of this in the work by Jaccard~\emph{et al.}~\cite{JaccardCars16,Jaccard2015} who found that log transforming images helped CNN-based ATD considerably.

Second, would ML-based material discrimination work better that the current physics-derived methods? ML methods might be better at exploiting spatial or contextual information to help in the presence of heavy noise found in commercial systems. With enough available data it might be possible to learn the material mapping using a fully Convolutional Neural Network~\cite{long2015fully}.

Third, do dual-energy systems actually aid automated Image Understanding? For example, can derived material information be used as a feature for ML algorithms? And can the $R$, $\alpha$, or H-L curves improve CNN approaches by being fed into the input channels? 

Fourth, the application of Deep Learning methods needs to be extended to Automated Contents Verification, in particular we feel they would be well suited to multi-class manifest verification.

Fifth, how do current and future algorithms compare to human operator performance? More work needs to be done on measuring baseline human performance, however there may be issues about disclosing these results to the public.

Finally, how transferable are currently developed algorithms - do they generalise to different scanning architectures? If not, can this be achieved through adequate data augmentation or transfer learning techniques?

\section*{Acknowledgement}
Funding for this work was provided through the EPSRC Grant no. EP/G037264/1 as part of UCL's Security Science Doctoral Training Centre, and Rapiscan Systems Ltd.

\bibliography{Review_paper,webSources}

\end{document}